\def\year{2020}\relax
\documentclass[letterpaper]{article} 
\usepackage{aaai20}  
\usepackage{times}  
\usepackage{helvet} 
\usepackage{courier}  
\usepackage[hyphens]{url}  
\usepackage{graphicx} 
\usepackage{graphicx}  
\nocopyright
\newcommand{\quotes}[1]{``#1''}

\usepackage{amsmath,amsfonts,bm}

\def\eqref#1{equation~\ref{#1}}

\def\1{\bm{1}}

\def\va{{\bm{a}}}

\def\vs{{\bm{s}}}

\DeclareMathAlphabet{\mathsfit}{\encodingdefault}{\sfdefault}{m}{sl}
\SetMathAlphabet{\mathsfit}{bold}{\encodingdefault}{\sfdefault}{bx}{n}

\def\sA{{\mathbb{A}}}

\def\sS{{\mathbb{S}}}

\newcommand{\E}{\mathbb{E}}

\newcommand{\R}{\mathbb{R}}

\DeclareMathOperator*{\argmin}{arg\,min}

\usepackage{algorithm}
\usepackage{algpseudocode}
\usepackage{pgfplots} 
\pgfplotsset{compat=1.14} 
\usepgfplotslibrary{statistics}
\usepgfplotslibrary{fillbetween}
\def \mujocoPlotWidth {0.65\columnwidth}
\def \mujocoPlotHeight {0.4\columnwidth}
\usepackage{subfigure}
\usetikzlibrary{external}
\usepackage{bibunits}
\defaultbibliography{qrrl}
\defaultbibliographystyle{aaai.bst}

\title{Learning Policies through Quantile Regression}
\author{ \Large \textbf{Oliver Richter and Roger Wattenhofer}\\ 
Departement of Electrical Engineering and Information Technology\\
ETH Zurich, Switzerland\\
\texttt{\{richtero,wattenhofer\}@ethz.ch} 
}
 
\begin{document}

\maketitle

\begin{abstract}
Policy gradient based reinforcement learning algorithms coupled with neural networks have shown success in learning complex policies in the model free continuous action space control setting. However, explicitly parameterized policies are limited by the scope of the chosen parametric probability distribution. We show that alternatively to the likelihood based policy gradient, a related objective can be optimized through advantage weighted quantile regression. Our approach models the policy implicitly in the network, which gives the agent the freedom to approximate any distribution in each action dimension, not limiting its capabilities to the commonly used unimodal Gaussian parameterization. This broader spectrum of policies makes our algorithm suitable for problems where Gaussian policies cannot fit the optimal policy. Moreover, our results on the MuJoCo physics simulator benchmarks are comparable or superior to state-of-the-art on-policy methods.
\end{abstract}

\section{Introduction}

Over the years, algorithms in artificial intelligence have become less restrictive in their assumptions, going from expert systems to kernel engineering to deep learning approaches, making the same algorithms applicable to a wider range of problems. In this mindset, we start from one of the least restrictive learning settings - continuous action space deep reinforcement learning - and show that removing the restriction of an explicitly parameterized policy can in fact improve performance. 
To achieve this, we cut loose of policy gradient methods and show that policies can be improved by moving probability mass through quantile regression towards adequate actions. 

The core idea of our algorithm is to use quantile regression~\cite{quantile_regression} for an implicit approximation of the optimal policy in the network parameters. Modeling the policy implicitly in the network allows us to approximate any distribution, i.e., we are not limited to an a-priori defined parameterized distribution (like multivariate Gaussian). This allows us to learn complex, multi-modal, non-Gaussian policies that are automatically inferred from interaction with the environment. While under full observability there exists an optimal deterministic policy~\cite{sutton} this is not the case in multi-agent and limited history setups. Here, stochastic, multi-modal policies are required to reach optimality; a complication that is often overlooked when applying an out-of-the-box Gaussian policy. On top of the stochastic flexibility we show that our algorithm has a trust region \cite{TRPO,PPO} interpretation and also achieves good performance on a large set of challenging control tasks from the OpenAI gym benchmark suite~\cite{gym}.

To summarize, our contributions are:
\begin{itemize}
    \item A derivation of a new reinforcement learning algorithm for continuous action space policies
    \item A discussion showing formal similarity between trust region policy gradient methods and quantile regression deep reinforcement policy learning
    \item An empirical evaluation showing the flexibility and competitiveness of the proposed algorithm
    \item And as a byproduct: A novel neural architecture for monotonic function approximation
\end{itemize}

\section{Background and Related Work}

In reinforcement learning~\cite{sutton} an \emph{agent} tries to learn a task through interaction with the environment in which the task is defined by means of a reward signal. More formally, in each time step $t$ the agent chooses an action $a_t$ based on the current state $\vs_t\in\sS$ and gets as feedback the reward $r_t$ and next state $\vs_{t+1}$ from the environment. $\sS$ denotes the set of all possible states. The goal of the agent is to maximize the discounted cumulative reward $R=\sum_t \gamma^t r_t$, with discount factor $\gamma\in [0,1]$, by adapting its policy to more rewarding trajectories. For our derivation, we will use the standard definitions \cite{sutton} of value function $V_\pi(\vs_t)$ and action-value function $Q_\pi(a_t,\vs_t)$
\[V_\pi(\vs_t) = \E_{\pi|\vs_t}\left[\sum_{t'=t}^\infty \gamma^{t'-t} r_{t'}\right]\]
\[Q_\pi(a_t, \vs_t) = \E_{\vs_{t+1}|a_t,\vs_t}[r_t + \gamma V_\pi(\vs_{t+1})]\]
where the expectation in $V_\pi$ is taken over trajectories collected when acting according to policy $\pi$ starting from state $\vs_t$. We 
use 
the star symbol (as in $\pi^*$) to denote optimality.

There are two main approaches in the literature to achieve reinforcement learning: Value based methods, foremost Q-learning~\cite{watkins1989learning}, and policy gradient based methods~\cite{sutton2000policy}.
Both methods have been adapted to use deep neural networks as function approximators~\cite{DQN,A3C} where policy gradient based methods are often combined with a critic into actor-critic methods to reduce the variance in the gradients. Deep reinforcement learning, the combination of deep learning and reinforcement learning, has also found its way into the continuous action space setting~\cite{DDPG}. Especially actor-critic methods with proximal second order updates~\cite{TRPO,ACKTR} - which ensure that the policy stays within a trust region - have shown success in learning complex policies in the continuous action space setting. A now popular algorithm, called \emph{Proximal Policy Optimization} (PPO), was introduced by \citeauthor{PPO}~\shortcite{PPO}, who showed that the trust region performance boost can be achieved by optimizing a clipped first order objective.

Recently, several papers \cite{energybased,soft1,q=ac,soft2,latent} showed that \emph{soft} Q-learning is in fact equivalent to maximum entropy actor-critic methods; a setup in which the agent is encouraged to maximize the entropy of its policy besides maximizing the reward. Combining a stochastic actor with the sample efficiency of Q-learning \cite{soft1,soft2} has led to impressive results. \citeauthor{latent}~\shortcite{latent} also extended this setup to a hierarchy of latent policies, where higher level policies control lower level actions through an invertible, normalizing flows network~\cite{normflow}. \citeauthor{boostnormflow}~\shortcite{boostnormflow} showed that a normalizing flows network can boost the performance of on-policy continuous action space algorithms as well. The drawback of normalizing flow networks is, however, that the network needs to be invertible and thereby requires the computation of the determinant of the Jacobian in each layer for the probability mass propagation. This limits normalizing flows in practice to narrow networks, a limitation not faced by quantile networks as in our setup.

One of the early works on maximum entropy deep reinforcement learning \cite{energybased} however already showed that one can train a particle based actor to approximate the (possibly multimodal) maximum-entropy policy defined by the soft Q-function with Stein-variational gradient descent~\cite{SVGD}. This work was extended by \citeauthor{WGF}~\shortcite{WGF} which showed that policy optimization can be seen as Wasserstein Gradient Flow in the policy probability space with the 2-Wasserstein distance as metric. This led them to propose an additional loss on the action particles which improved performance. The relationship between policy optimization and Wasserstein Gradient Flows was simultaneously shown by \citeauthor{WGF_}~\shortcite{WGF_}. The problem with particle based methods is, however, that for a general sampling network the likelihood of a given sample cannot be directly recovered. Therefore, off-policy corrections for multi-step learning as proposed in \cite{Retrace,impala} cannot be applied. As we discuss 
in Appendix~\ref{app:likelihood}, the sample likelihood can be retrieved from the quantile function. Our method could therefore in principle be extended with these off-policy corrections in future work.

\citeauthor{VIREL}~\shortcite{VIREL} recently introduced a variational inference framework for deep reinforcement learning. They investigated the probabilistic nature of policies learned by deep reinforcement learning algorithms from a variational inference perspective, but revert to Gaussian policies in their experiments for implementation convenience.

\citeauthor{QR-DQN}~\shortcite{QR-DQN,IQN} were the first to use quantile regression in connection with deep reinforcement learning. In their work,
they focused on
approximating the full probability distribution of the value function. In contrast, we
explore possibilities of using quantile regression to approximate richer policies by not constraining the action distribution to an explicitly parameterized distribution. \citeauthor{QN_generative_modeling}~\shortcite{QN_generative_modeling} showed that quantile networks can also be used for generative modeling while recently \citeauthor{quantile_forecasting}~\shortcite{quantile_forecasting} used quantile regression in combination with a recurrent neural network for probabilistic forecasting. In general, we see quantile regression in combination with deep learning to have a lot of potential for future work as it provides a flexible way to represent an arbitrary probability distribution within the network.

\section{Quantile Regression and Quantile Networks}
\label{quant_net}

Quantile regression \cite{quantile_regression} discusses approximation techniques for the inverse cumulative distribution function $F_Y^{-1}$, i.e., the \emph{quantile function}, of some probability distribution $Y$. Recent work \cite{IQN,QN_generative_modeling} showed that a neural network can learn to approximate the quantile function by mapping a uniformly sampled quantile target $\tau \sim \mathcal{U}([0,1])$ to its corresponding quantile function value $F_Y^{-1}(\tau)\in \R$. Thereby, the trained neural network implicitly models the full probability distribution $Y$. 

More formally, let $W_p(U,Y)$ be the p-Wasserstein metric
\[W_p(U,Y) = \left(\int_0^1 |F_Y^{-1}(\omega) - F_U^{-1}(\omega)|^p d\omega\right)^{1/p}\]
of distributions $U$ and $Y$, also characterized as the $L^p$ metric of quantile functions \cite{muller1997integral}. \citeauthor{QR-DQN}~\shortcite{QR-DQN} show that minimizing the quantile regression loss \cite{quantile_regression_loss}
\begin{equation}
\label{quantile_loss}
    \rho_\tau(\delta) = (\tau - \1_{\delta < 0})\cdot\delta 
\end{equation}

reduces the 1-Wasserstein distance between a scalar target probability distribution $Y$ and a parameterized mixture of Diracs $U_\theta$, which correspond to fixed quantiles $\tau$. Here, $\delta = y - u$ with $y\sim Y$ and $u = F_U^{-1}(\tau)$ is the quantile sample error. \citeauthor{QN_generative_modeling}~\shortcite{QN_generative_modeling} generalized this result by showing that the expected quantile loss
\begin{equation}
    \label{expected_quantile_loss}
    \mathcal{L} = \E_{\tau\sim \mathcal{U}([0,1])} \left[\E_{z\sim Z}\left[\rho_\tau(z - \hat{G}_\theta(\tau))\right]\right]
\end{equation}
of a parameterized quantile function $\hat{G}_\theta$ aproximating the quantile function $F_Z^{-1}$ of some distribution $Z$ is equal to the quantile divergence

\[
    q(Z,\pi_\theta) := \int_0^1\left[\int_{F_Z^{-1}(\tau)}^{F_{\pi_\theta}^{-1}(\tau)}(F_Z(x)-\tau)dx\right]d\tau
\]

plus some constant not depending on the parameters $\theta$. Here, $\pi_\theta$ is the distribution implicitly defined by $\hat{G}_\theta$. Therefore, training a neural network $\hat{G}_\theta(\tau)$ to minimize $\rho_\tau(z-\hat{G}_\theta(\tau))$ with $z$ sampled from the target probability distribution $Z$ effectively minimizes the quantile divergence $q(Z,\pi_\theta)$ and thereby models an approximate distribution $\pi_\theta$ of $Z$ implicitly in the network parameters $\theta$ of the neural network $\hat{G}_\theta(\tau)$. 

Another way to state this result is by noting that the quantile regression loss $\rho_\tau$ appears in the \emph{continuous ranked probability score} (CRPS)~\cite{10.2307/2629907}
\[CRPS(F^{-1}, z) := \int_0^1 2 \rho_\tau(z - F^{-1}(\tau)) d\tau  \]
which is a proper scoring rule \cite{doi:10.1198/016214506000001437}, i.e., $\E_{z\sim Z}[CRPS(F_Z^{-1}, z)] \leq \E_{z\sim Z}[CRPS(F_Y^{-1}, z)]$ for any distributions $Z$ and $Y$. Minimizing the expected quantile loss (\ref{expected_quantile_loss}) is equivalent to minimizing the expected score $\E_{z\sim Z}[CRPS(\hat{G}_\theta, z)]$ which leads to $\hat{G}_\theta$ approximating $F_Z^{-1}$ (derivation adapted from \citeauthor{quantile_forecasting}~\shortcite{quantile_forecasting}). Note that this derivation requires $\hat{G}_\theta$ to always define a proper quantile function, which we ensure in this work by modeling $\hat{G}_\theta$ through a monotonically non-decreasing neural network.

By approximating the quantile function instead of a parameterized probability distribution, as it is common in many deep learning models \cite{VAE,A3C,TRPO,ACKTR,PPO}, we do not enforce any constraint on the probability distribution $Z$. E.g., $Z$ can be multi-modal, i.e., non-Gaussian, as a strong function approximator, like a neural network, can approximate the corresponding quantile function.

\section{Quantile Regression Reinforcement Learning}\label{qrrl} 

Given that a quantile network can approximate any probability distribution, we aim at approximating the optimal policy in a reinforcement learning setup. For this, we model for each action dimension the quantile function $\hat{G}_\theta(\tau, \vs) = F_{\pi_\theta}^{-1}(\tau, \vs)$ of the implicitly defined action distribution $\pi_\theta$ by a neural network with a state $\vs$ and a target quantile $\tau\in [0,1]$ as input. In practice, we share the first few layers extracting features from the state into an embedding $\phi(\vs)$, before we pass the embedding to the individual quantile networks $\hat{G}_\theta(\tau, \phi(\vs))$. However, for ease of notation we omit this shared network $\phi(\cdot)$ in the following derivations.\footnote{See Appendix~\ref{app:RL} for more details. Our exact implementation can be found in the supplementary material.} From the full network $\bm{\hat{G}}_\theta: [0,1]^d\times \sS\to\R^d$, with $d$ being the number of action dimensions, an action $\va\in\sA\equiv\R^d$ for a given state $\vs$ can be obtained by sampling $\bm{\tau}\sim\mathcal{U}([0,1]^d)$ and taking the network output as action. Since the network approximates quantile functions, the network output of a uniformly at random sampled quantile target is a sample from the implicitly defined action distribution. The question left to address is how to train the network, such that it (a) represents the quantile functions of the action dimensions and (b) the implicitly defined policy maximizes the expected (discounted) reward $R$.

Objective (a) can be achieved by limiting the quantile network to a monotonic function with respect to the quantile input $\tau$. This ensures that the network represents a valid quantile function already at the start of training. Note that such an architectural prior, although sensible, was not applied in related work~\cite{IQN,QN_generative_modeling}. Here however, it is vital, since it allows us to perform gradient ascent on the quantile loss
(detailed below), as the network remains a valid representation of a quantile function. A monotonically non-decreasing neural network can be achieved by restricting the neuron connecting weights to be positive and applying a non-linearity that allows for concavity as well as convexity. For weight positivity we simply set the network weights $\theta^W_{net}$ of the quantile net to the element-wise exponent of unconstrained weights $\theta^W$, i.e., $\theta^W_{net} = \exp{\theta^W}$. As non-linearity, we choose a combination of ReLUs, with neuron output $y=\max(0,x)$ given neuron input $x$, and inverse ReLUs, where $y=\min(0,x)$.  We detail and contrast our architecture design against other designs in
Appendix~\ref{app:monotonic_net}, extending the current literature on monotonic networks.

The more difficult objective is (b); achieving a policy improvement over time. Here we address this with advantage weighted quantile regression. Informally put, quantile regression is linked to the Wasserstein metric which is also sometimes referred to as \emph{earth mover's distance}. Imagine a pile of earth representing probability mass. In reinforcement learning we essentially want to move probability mass towards actions that were good and away from actions that were bad, where \quotes{good} and \quotes{bad} are measured by discounted cumulative (bootstrapped) reward received. Quantile regression can achieve this
by shaping the pile of earth according to an advantage estimation and the constraint of monotonicity
(a core property of quantile functions).

More formally, we are interested in approximating the optimal policy $\pi^*(a)$, where we omit the implicit dependence on the state $\vs$ in the following for ease of readability.\footnote{We focus our analysis on the simple case of a single action dimension with scalar input and scalar output. The generalization to multidimensional independent action-/quantile-functions follows trivially.}
If we were given the optimal policy, this could be achieved by training on the quantile regression objective
\[\theta^* = \argmin_\theta \E_{a\sim\pi^*}\left[\E_{\tau\sim\mathcal{U}([0,1])}\left[\rho_\tau(a - \hat{G}_\theta(\tau))\right]\right] \]
as this would minimize the quantile divergence between $\pi^*$ and $\pi_\theta$ as derived in the previous section. However, sampling from the optimal policy $\pi^*$ is infeasible, since we do not know it a-priori.
Therefore we rewrite our objective as
\[\theta^* = \argmin_\theta\E_{a\sim \mu}\left[\E_{\tau\sim\mathcal{U}([0,1])}\left[\frac{\pi^*(a)}{\mu(a)}\rho_\tau(a - \hat{G}_\theta(\tau))\right]\right] \]
where $\mu$ is a policy with support greater or equal to the support of $\pi^*$. The importance ratio $\frac{\pi^*(a)}{\mu(a)}$ gives a measure of how much more/less likely a given action $a$ would be under the optimal policy $\pi^*$ compared to the policy $\mu$ that collected the experience. 
To ease our calculations, we assume the behaviour policy $\mu$ to be uniform, i.e., to have constant likelihood $\mu(a)$ for all actions $a$ within the support of $\mu$. This lets us absorb the experience likelihood $\mu(a)$ into the optimization.
Appendix~\ref{app:constant_mu} discusses this approximation further.

Following~\cite{VIREL,energybased} we define the optimal soft policy at temperature $\beta$ as $\pi_\beta^*(a)\propto\exp{\left(\frac{Q^*(a) - V^*}{\beta}\right)}$, where $Q^*$ and $V^*$ are the soft action-value and value function of the optimal policy. We choose the soft formulation to retain a stochastic policy, as we are interested in setups, where the optimal policy is not deterministic.
We can then write our objective as
\begin{equation*}
    \theta^* = \argmin_\theta \E_{a\sim \mu}\left[\E_{\tau\sim\mathcal{U}([0,1])}\left[\mathcal{L}_\theta(a,\tau)\right]\right]
\end{equation*}
\[\text{with}\quad\mathcal{L}_\theta(a,\tau) = \exp\left(
    \frac{Q^*(a) - V^*}{\beta}\right)\rho_\tau(a - \hat{G}_\theta(\tau))\]
This shifts the perspective from estimating the likelihood under the optimal policy to estimating the advantage $A^*(a) = Q^*(a) - V^*$ of the optimal policy over the (sub-optimally) chosen action $a\sim\mu$. While this quantity is still unknown a-priori, we can approximate the advantage of our current policy $\pi_\theta$. If we therefore approximate the optimal soft policy by the exponential of $A_{\pi_\theta}$, the advantage of our current best policy, we arrive at the iterative optimization procedure

\begin{equation}\label{contraction_objective}
    \theta_{k+1} = \argmin_\theta \E_{a\sim \mu}\left[\E_{\tau\sim\mathcal{U}([0,1])}\left[\mathcal{L}_{\theta}^k(a,\tau)\right]\right]
\end{equation}
\[\text{with}\quad\mathcal{L}_{\theta}^k(a,\tau) = \exp\left(
    \frac{A_{\theta_k}(a)}{\beta}\right)\rho_\tau(a - \hat{G}_\theta(\tau))\]
where we used the short notation $A_{\theta_k}$ to denote the advantage of policy $\pi_{\theta_k}$.
Note that this objective has a (stochastic) convergence fixpoint with $\pi_\theta \equiv \pi_\beta^*$.
In practice however, it is cumbersome to define a behaviour policy $\mu$ for experience collection a-priori, because an a-priori defined $\mu$ might not cover the support of the optimal policy $\pi_\beta^*$, as required by the derivation. On the other hand, an a-priori defined $\mu$ that does cover the support of $\pi_\beta^*$ might be too explorative, rendering only a few informative action samples, which would result in a low sample efficiency of the algorithm. What we would therefore like to do is to use our current best estimate of the optimal policy to gather experience. Note, however, that the optimization~(\ref{contraction_objective}) contracts $\pi_\theta$ onto the support of $\mu$, since the objective regresses $\pi_\theta$ towards samples of $\mu$. If we set $\mu\equiv\pi_\theta$ and use an approximate iterative procedure like stochastic gradient descent, at some point the support of $\pi_\theta$ would not cover the support of $\pi^*$ anymore and the policy would degenerate in the limit towards a sub-optimal deterministic action in each state. Heuristics, such as entropy regularization~\cite{A3C}, have been shown to circumvent similar problems in the discrete action space setup. Note however, that entropy regularization would be non-trivial in our quantile function continuous action space setup. Instead, we propose the following: we take a linear approximation to the exponential function $\exp(x)\approx x + 1$ around $x\approx0$ and multiply by $\beta$ to get the regularized iterative objective function
\begin{equation}\label{final_objective}
\mathcal{L}_{\theta}^k(a,\tau) \approx A_{\theta_k}(a)\rho_\tau(a - \hat{G}_\theta(\tau)) + \beta\rho_\tau(a - \hat{G}_\theta(\tau))
\end{equation}
Note that this approximation is reasonable for $A_{\theta_k}(a)\approx 0$, which in turn is reasonable for $a\sim\pi_{\theta_k}$ given a decent approximation of the advantage. This linearization however brings an interesting property: if an action taken by the behaviour policy results in an outcome that is worse than the outcome predicted for $\pi_{\theta_k}$, i.e., the advantage $A_{\theta_k}(a)$ is negative, the objective regularizes in that we maximize the corresponding quantile loss $\rho_\tau$. This essentially pushes action samples $a'\sim\pi_\theta$ away from the bad action $a$, thereby expanding the support of $\pi_\theta$. Note that ascending on the quantile loss does not lead to divergence since we restrict our quantile network to be monotonically non-decreasing
and therefore any parametarization of the network results in a valid quantile function. 
We can therefore now safely replace the behaviour policy $\mu$ to generate experience with our current best estimate $\pi_{\theta_k}$. This gives us an effective, iterative on-policy algorithm, that regresses towards the optimal soft policy based on experience, while regularizing against a premature collapse to a local optimum.

Another motivation for objective~(\ref{final_objective}) is its similarity to trust region methods \cite{TRPO,PPO}: while the first part in objective~(\ref{final_objective}), $A_{\theta_k}(a)\rho_\tau(a - \hat{G}_\theta(\tau))$, gives a measure for how much we will update our policy based on the current rollout, the second part, $\beta\rho_\tau(a - \hat{G}_\theta(\tau))$, constrains the updated policy to not deviate too far from the behaviour policy which gathered the experience. I.e., the second term defines a trust region objective, which is controlled by $\beta$. A high value for $\beta$ keeps the policy close to the policy that collected the current batch of experience, yielding update stability, while a low value for $\beta$ allows the policy to adjust more rapidly to new experiences. We use this similarity of our algorithm to trust region methods to take advantage of algorithmic improvements employed by \citeauthor{PPO}~\shortcite{PPO} and adapt them without adjustment in our algorithm. Namely, in our reinforcement learning experiments we use generalized advantage estimation~\cite{GAE} to approximate $A_{\theta_k}(a)$, normalize the advantages 
and train a small number of epochs with mini-batches on each collected rollout. Pseudo code is available in Appendix~\ref{app:pseudocode}.

Further, note that the number of samples taken from the inner expectation in objective~(\ref{contraction_objective}) can be 
chosen to trade-off required computation against gradient variance. In some initial experiments on the MuJoCo Swimmer and Ant task we varied the number of samples $K\in \{1,32,128,256\}$ and chose $K=128$ as we found this to be a good trade-off. Similarly we chose $\beta=2$ from $\{0,1,2,4\}$ based on the same tasks. The remaining hyper-parameters were copied without adjustment from \citeauthor{PPO}~\shortcite{PPO} and are reported in Appendix~\ref{app:RL}. The fact that our algorithm works without any adjustment of these hyper-parameters hints at its robustness and ease of applicability. We leave it to future work to find a better 
hyper-parameter set.


\begin{figure}[t]
\centering
\subfigure{
\begin{tikzpicture}
\begin{axis}[
    ybar,
    ymin=0,
    axis lines=left,
    hide y axis,
    width=1.1\columnwidth,
    height=0.3\columnwidth
]
\addplot +[
    hist={
        bins=50,
        data min=-1.6,
        data max=1.6
    },
] table [y index=0] {data/rock_paper_scissors/qrrl_actions_small.csv};
\end{axis}
\tikzexternalize[prefix=tikz/]
\end{tikzpicture}}
\subfigure{
\centering
\includegraphics[width=0.95\columnwidth]{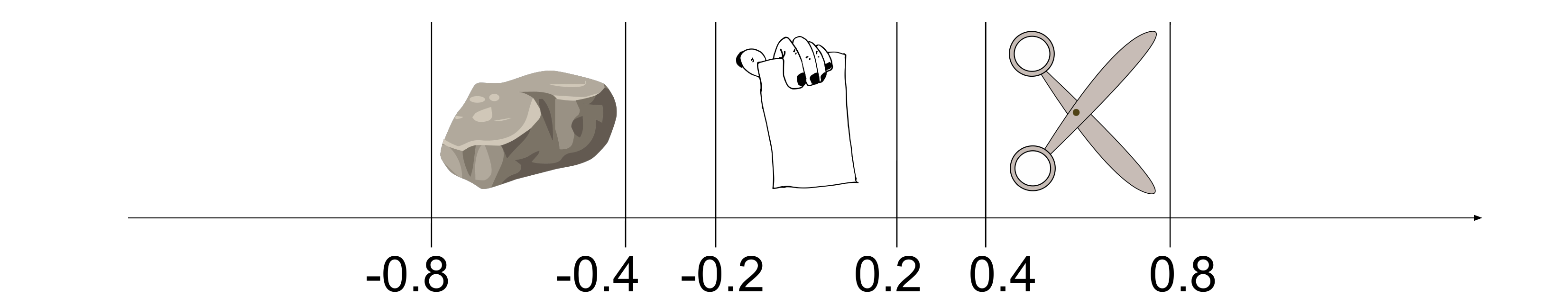}}
\subfigure{
\begin{tikzpicture}
\begin{axis}[
    ybar,
    ymin=0,
    axis lines=left,
    hide y axis,
    width=1.1\columnwidth,
    height=0.3\columnwidth
]
\addplot +[
    hist={
        bins=50,
        data min=-1.6,
        data max=1.6
    },
    color=green!50.19607843137255!black,
    fill=green!50.19607843137255!black!20,
] table [y index=0] {data/rock_paper_scissors/pg_actions_small.csv};
\end{axis}
\tikzexternalize[prefix=tikz/]
\end{tikzpicture}}
\caption{\textbf{Middle:} Decomposition of the 1-dimensional continuous action space into the three distinct actions of the rock paper scissors game. Any action outside the indicated intervals is treated as an invalid action and results in a loss if the opponent chooses a valid action. \textbf{Top:} Action distribution learned by quantile regression training. \textbf{Bottom:} Action distribution learned by policy gradient training.}
\label{fig:rps_action_space}
\end{figure}
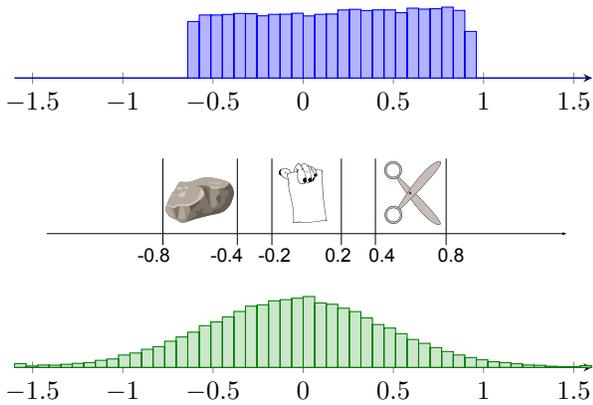
\begin{figure}[t]
\centering
	\begin{tikzpicture}
	\begin{axis}[width = \columnwidth, height = 0.72\columnwidth, legend pos=north west, legend style={cells={align=center}}]
	\addplot[blue, thick, mark=] table [col sep=comma] {data/rock_paper_scissors/qrrlav.csv};
	\addplot[green!50.19607843137255!black, thick, mark=] table [col sep=comma] {data/rock_paper_scissors/pgav.csv};
	\legend{Quantile Net, Gaussian Net};
	\addplot[green!50.19607843137255!black, opacity=0, name path=pg_upper, mark=] table [col sep=comma] {data/rock_paper_scissors/pgup.csv};
	\addplot[green!50.19607843137255!black, opacity=0, name path=pg_lower, mark=] table [col sep=comma] {data/rock_paper_scissors/pglo.csv};
	\addplot[green!50.19607843137255!black, fill opacity=0.2] fill between[of=pg_upper and pg_lower];
	\addplot[blue, opacity=0, name path=qrrl_upper, mark=] table [col sep=comma] {data/rock_paper_scissors/qrrlup.csv};
	\addplot[blue, opacity=0, name path=qrrl_lower, mark=] table [col sep=comma] {data/rock_paper_scissors/qrrllo.csv};
	\addplot[blue, fill opacity=0.2] fill between[of=qrrl_upper and qrrl_lower];
	\end{axis}
	\end{tikzpicture}
	\caption{Average return of the two policy network types over the course of training rock-paper-scissors (see main text for details). The x-axis denotes the training iteration
	. A return of -1 corresponds to the countering policy always winning while a return of 1 corresponds to the training policy always winning. Plotted is the average and standard deviation (shaded area) of 20 independent runs. Curves are smoothed over 10 training iterations, i.e., 1000 games.}
	\label{fig:rock_paper_scissors}
\end{figure}
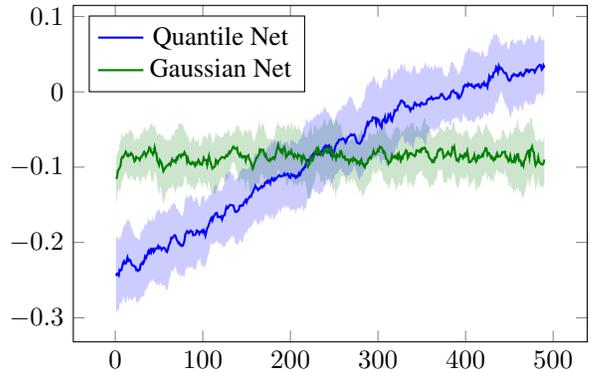

\section{Experiments and Results}

As recent papers have raised valid concerns about the reproducibility of deep reinforcement learning results in the continuous action domain \cite{reproducible,drl_that_matters}, we ran all our experiments for 20 fixed random seeds (0-19). Implementation details and hyper-parameter choices to reproduce the results can be found in the supplementary material. With our experiments we aim to answer the following questions:

\begin{itemize}
    \item Is the ability to learn more diverse probability distributions worthwhile to get an advantage in simple two player games? Do multimodal policies emerge from training on these games?
    \item How does quantile regression deep reinforcement learning (or QRDRL in short) compare to other online algorithms on well studied reinforcement learning benchmarks?
\end{itemize}

As a first experiment, we implemented a continuous action version of Rock-Paper-Scissors. Rock-Paper-Scissors can be seen as a two player multi-arm bandit problem, each player chooses one of the actions \quotes{Rock}, \quotes{Paper} or \quotes{Scissors} and the winner is determined based on the choices: Rock beats Scissors, Scissors beats Paper and Paper beats Rock. We modeled the players through algorithms choosing their action in a 1-dimensional continuous action space, where we defined intervals for the corresponding discrete actions as shown in Figure~\ref{fig:rps_action_space}. We aim to learn a non-exploitable policy in this setup, that is, we train a policy such that a countering policy, trained to exploit the former policy, achieves the minimum possible wins. Specifically, in each training iteration we train a countering Gaussian policy from scratch on 10,000 games against the current policy and then use this countering policy as opponent in 100 games based on which the current policy is updated.

We trained two policy networks in this setup: (i) a quantile network trained on the weighted quantile loss $\E_{a\sim \pi_\theta}[\E_{\tau\sim\mathcal{U}([0,1])}[r\cdot \rho_\tau(a - \hat{G}_\theta(\tau))]]$, where the weight $r=1$ for games that were won and $r=-1$ for games that were lost,\footnote{This is the multi-arm bandit analogy to (\ref{final_objective}) with $\beta=0$.} and (ii) a Gaussian policy trained to maximize the log-likelihood of winning, i.e., maximize $r$, via policy gradient. The exact experiment setup is described in Appendix~\ref{app:rps}.
Figure~\ref{fig:rps_action_space} shows histograms of the action distributions learned by the two approaches while Figure~\ref{fig:rock_paper_scissors} shows the average return $r$ throughout the training. The results show that the uni-modal nature of the Gaussian network can always be exploited by the countering policy, hindering any learning progress. On the other hand, the Quantile network learns to choose close to uniform at random, making the policy impossible to exploit. Moreover, it learned that the countering Gaussian policy is initialized with the mean close to 0 - \quotes{Paper} - which explains the slight tilt of the action distribution towards the right - \quotes{Scissors} - and the slightly above zero return at the end of the training. I.e., it has learned to exploit the initialization and inability to counter of the countering policy.

However, we did not find the multi-modal nature in the learned action distribution that we were hoping for. To verify that our approach can indeed learn a multi-modal policy we implemented another toy game. In this game, which we call Choice, the agent also acts in a single continuous action dimension, where an action between -0.6 and -0.4 corresponds to a button A pressed while an action between 0.4 and 0.6 corresponds to a button B pressed. The agent is rewarded if it presses the button that was pressed less often so far within the episode. The problem is complicated in that we only model the agent as a feed-forward network, giving it no ability to remember the actions it took so far. Note that in this setup the best policy is to choose one of the two buttons at random. Implementation details can be found in Appendix~\ref{app:RL}.
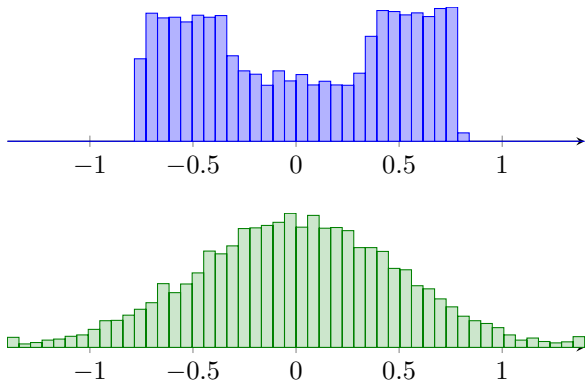
\begin{figure}[t]
\centering
\subfigure{
\begin{tikzpicture}
\begin{axis}[
    ybar,
    ymin=0,
    axis lines=left,
    hide y axis,
    width=1.1\columnwidth,
    height=0.4\columnwidth
]
\addplot +[
    hist={
        bins=50,
        data min=-1.4,
        data max=1.4
    },
] table [y index=0] {data/choice/qrrl_actions.csv};
\end{axis}
\tikzexternalize[prefix=tikz/]
\end{tikzpicture}}
\subfigure{
\begin{tikzpicture}
\begin{axis}[
    ybar,
    ymin=0,
    axis lines=left,
    hide y axis,
    width=1.1\columnwidth,
    height=0.4\columnwidth
]
\addplot +[
    hist={
        bins=50,
        data min=-1.4,
        data max=1.4
    },
    color=green!50.19607843137255!black,
    fill=green!50.19607843137255!black!20,   
] table [y index=0] {data/choice/ppo_actions.csv};
\end{axis}
\tikzexternalize[prefix=tikz/]
\end{tikzpicture}}
\caption{Histograms of the learned action distributions on the choice toy game. \textbf{Top:} Our approach (QRDRL) \textbf{Bottom:} PPO with gaussian policy}
\label{fig:choice_actions}
\end{figure}
\begin{figure}[t]
\centering
	\input{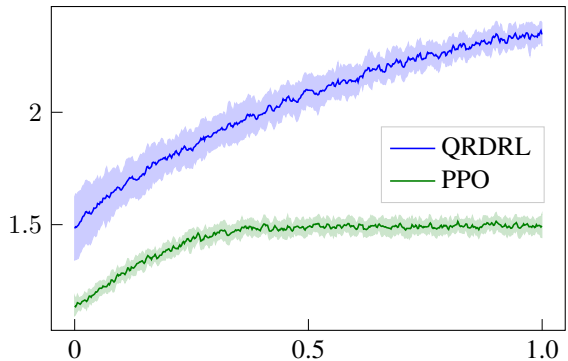}
	\caption{Average return over the course of training on the choice toy game. The x-axis denotes steps in millions. Plotted is the average and standard deviation (shaded area) of 20 independent runs.}
	\label{fig:choice_return}
\end{figure}

As can be seen in Figure~\ref{fig:choice_actions}, QRDRL is capable of recovering the two modes needed to solve the task while proximal policy optimization (PPO, \citeauthor{PPO}~\shortcite{PPO}), a commonly used Gaussian policy gradient method, learns a suboptimal compromise between the two buttons. This is especially apparent when we look at the corresponding return throughout the training depicted in Figure~\ref{fig:choice_return}: the return of PPO stagnates around 1.5 while QRDRL continues to improve throughout the training. Nevertheless, QRDRL was unable to put 0 probability on the invalid actions between the modes within the given training time. We believe this stems mainly from our architecture choice and the artificial discontinuous distribution setup that is difficult to approximate (see Appendix~\ref{app:monotonic_net}). 

Given that the ability to express more complex stochastic policies is indeed vital to perform well in the toy games presented, we are left with our second research question, whether QRDRL also performs well on commonly used reinforcement learning benchmarks. To this end, we run our algorithm on a diverse set of robotic tasks defined in OpenAI gym~\cite{gym} based on the MuJoCo~\cite{mujoco} physics simulator. These include diverse robotic figures which should learn to walk as well as robotic arms and pendulums which need to reach a certain point or balance themselves. We compare our approach against PPO~\cite{PPO} as well as normalizing flows TRPO~\cite{normflow}, a recently proposed on-policy algorithm using normalizing flows.\footnote{We compare to their TRPO version since their normalizing flow ACKTR implementation yielded worse/inconsistent results.} Implementation details are provided in Appendix~\ref{app:RL}

\begin{figure*}
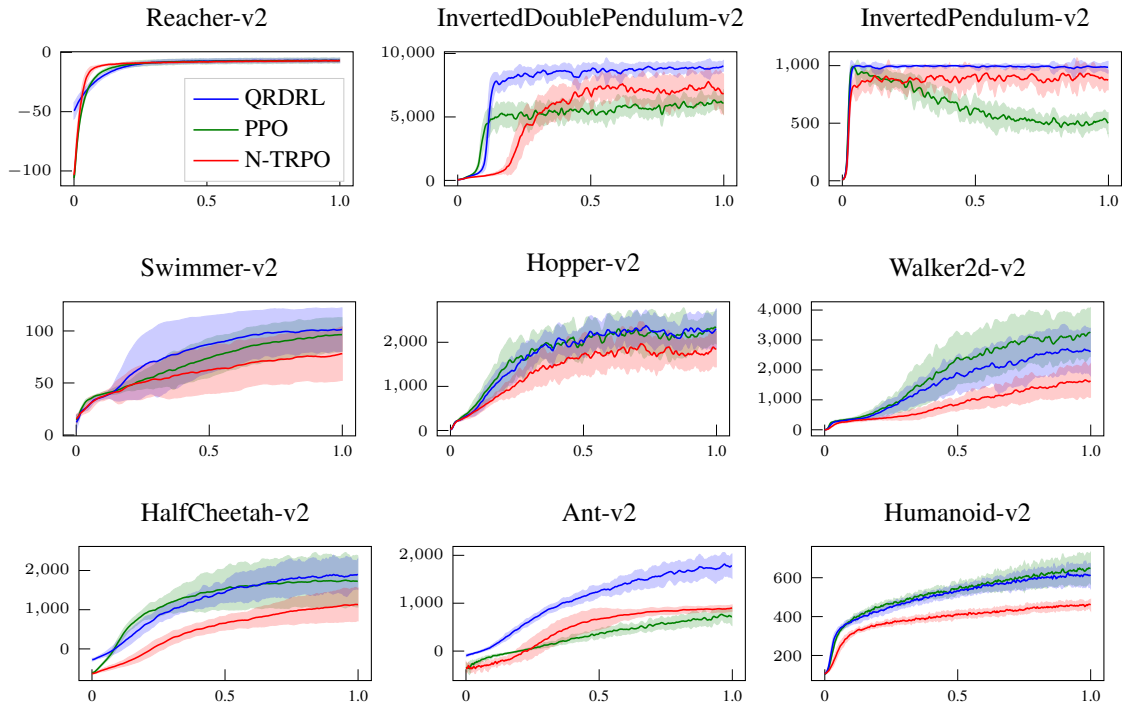

\centering
    \subfigure{\input{figures/mujoco/Reacher_plot.tex}}
    \subfigure{\input{figures/mujoco/InvertedDoublePendulum_plot.tex}}
    \subfigure{\input{figures/mujoco/InvertedPendulum_plot.tex}}
    \subfigure{\input{figures/mujoco/Swimmer_plot.tex}}
    \subfigure{\input{figures/mujoco/Hopper_plot.tex}}
    \subfigure{\input{figures/mujoco/Walker2d_plot.tex}}
    \subfigure{\input{figures/mujoco/HalfCheetah_plot.tex}}
    \subfigure{\input{figures/mujoco/Ant_plot.tex}}
    \subfigure{\input{figures/mujoco/Humanoid_plot.tex}}
    \caption{Average return on OpenAI gym MuJoCo benchmarks. The x-axis denotes steps in millions. Plotted is the average and standard deviation (shaded area) of 20 independent runs.} 
    \label{fig:mujoco}
\end{figure*}

The results reported in Figure~\ref{fig:mujoco} show that QRDRL is indeed capable of performing as well as, if not superior to current state-of-the-art on-policy methods. Specifically, we clearly outperform these methods on the pendulum tasks as well as the challenging Ant task. We see a similar trend in the normalizing flows TRPO (N-TRPO) results, suggesting that a more flexible policy is especially helpful in these tasks. However, N-TRPO fails to match our results in many cases, which we believe might be due to the limitation of a narrow flow network. Note also that we achieved our results without extensive hyper-parameter tuning, and our results might even improve for better suited hyper-parameters. 

\section{Conclusion and Outlook}

In this work, we introduce quantile regression deep reinforcement learning (QRDRL), a likelihood free reinforcement learning approach to overcome the limitation of explicitly parameterized policies. We show that the ability to learn more diverse stochastic continuous action space policies is essential to perform well in different situations, be it to defer the exploitability of a policy or to cope with memory limitations. We further show that our algorithm has performance comparable to state-of-the-art on-policy methods on the well studied MuJoCo benchmarks.

Given that this work introduces a new idea to approach reinforcement learning with quantile regression, it offers many directions for future work:

\textbf{Off-policy learning:} We showed in this work how quantile regression can be used within an on-policy algorithm. Future work could extend our approach to the off-policy setting, leveraging the advantage of sample reuse. Note that while our training objective is likelihood free, we can still recover the action likelihood through a simple gradient back propagation (see Appendix~\ref{app:likelihood}). This is a distinct advantage of our approach over other particle based algorithms, as it enables the potential use of multi-step off policy corrections as described by \citeauthor{Retrace}~\shortcite{Retrace} or \citeauthor{impala}~\shortcite{impala}.

\textbf{Multivariate quantile regression:} We took a conservative approach in this work by modeling each action dimension as an independent quantile function. This is along the line of Gaussian policy implementations, where off-diagonal elements of the covariance matrix are omitted as well. However, insights into multivariate quantile regression \cite{multivariate_qr,multivariate_qr2} could be adapted to give our algorithm the capability to infer the stochastic relations between action dimensions.



\bibliographystyle{aaai.bst}
\bibliography{qrrl}

\begin{thebibliography}{}

\bibitem[\protect\citeauthoryear{Brockman \bgroup et al\mbox.\egroup
  }{2016}]{gym}
Brockman, G.; Cheung, V.; Pettersson, L.; Schneider, J.; Schulman, J.; Tang,
  J.; and Zaremba, W.
\newblock 2016.
\newblock Openai gym.

\bibitem[\protect\citeauthoryear{Cano \bgroup et al\mbox.\egroup
  }{2019}]{monotonic_overview}
Cano, J.-R.; Gutiérrez, P.~A.; Krawczyk, B.; Woźniak, M.; and García, S.
\newblock 2019.
\newblock Monotonic classification: An overview on algorithms, performance
  measures and data sets.
\newblock {\em Neurocomputing} 341:168 -- 182.

\bibitem[\protect\citeauthoryear{Chakraborty}{2003}]{multivariate_qr}
Chakraborty, B.
\newblock 2003.
\newblock On multivariate quantile regression.
\newblock {\em Journal of Statistical Planning and Inference} 110(1):109 --
  132.

\bibitem[\protect\citeauthoryear{Dabney \bgroup et al\mbox.\egroup
  }{2018a}]{IQN}
Dabney, W.; Ostrovski, G.; Silver, D.; and Munos, R.
\newblock 2018a.
\newblock Implicit quantile networks for distributional reinforcement learning.
\newblock In {\em Proceedings of the 35th International Conference on Machine
  Learning, {ICML} 2018, Stockholmsm{\"{a}}ssan, Stockholm, Sweden, July 10-15,
  2018},  1104--1113.

\bibitem[\protect\citeauthoryear{Dabney \bgroup et al\mbox.\egroup
  }{2018b}]{QR-DQN}
Dabney, W.; Rowland, M.; Bellemare, M.~G.; and Munos, R.
\newblock 2018b.
\newblock Distributional reinforcement learning with quantile regression.
\newblock In {\em Proceedings of the Thirty-Second {AAAI} Conference on
  Artificial Intelligence, New Orleans, Louisiana, USA, February 2-7, 2018}.

\bibitem[\protect\citeauthoryear{{Daniels} and
  {Velikova}}{2010}]{overview_mono_net_2}
{Daniels}, H., and {Velikova}, M.
\newblock 2010.
\newblock Monotone and partially monotone neural networks.
\newblock {\em IEEE Transactions on Neural Networks} 21(6):906--917.

\bibitem[\protect\citeauthoryear{Dinh, Sohl{-}Dickstein, and
  Bengio}{2017}]{normflow}
Dinh, L.; Sohl{-}Dickstein, J.; and Bengio, S.
\newblock 2017.
\newblock Density estimation using real {NVP}.
\newblock In {\em 5th International Conference on Learning Representations,
  {ICLR} 2017, Toulon, France, April 24-26, 2017, Conference Track
  Proceedings}.

\bibitem[\protect\citeauthoryear{Espeholt \bgroup et al\mbox.\egroup
  }{2018}]{impala}
Espeholt, L.; Soyer, H.; Munos, R.; Simonyan, K.; Mnih, V.; Ward, T.; Doron,
  Y.; Firoiu, V.; Harley, T.; Dunning, I.; Legg, S.; and Kavukcuoglu, K.
\newblock 2018.
\newblock {IMPALA:} scalable distributed deep-rl with importance weighted
  actor-learner architectures.
\newblock In {\em Proceedings of the 35th International Conference on Machine
  Learning, {ICML} 2018, Stockholmsm{\"{a}}ssan, Stockholm, Sweden, July 10-15,
  2018},  1406--1415.

\bibitem[\protect\citeauthoryear{Fellows \bgroup et al\mbox.\egroup
  }{2018}]{VIREL}
Fellows, M.; Mahajan, A.; Rudner, T. G.~J.; and Whiteson, S.
\newblock 2018.
\newblock {VIREL:} {A} variational inference framework for reinforcement
  learning.
\newblock {\em CoRR} abs/1811.01132.

\bibitem[\protect\citeauthoryear{Gasthaus \bgroup et al\mbox.\egroup
  }{2019}]{quantile_forecasting}
Gasthaus, J.; Benidis, K.; Wang, Y.; Rangapuram, S.~S.; Salinas, D.; Flunkert,
  V.; and Januschowski, T.
\newblock 2019.
\newblock Probabilistic forecasting with spline quantile function rnns.
\newblock In Chaudhuri, K., and Sugiyama, M., eds., {\em Proceedings of Machine
  Learning Research}, volume~89 of {\em Proceedings of Machine Learning
  Research},  1901--1910.
\newblock PMLR.

\bibitem[\protect\citeauthoryear{Glorot, Bordes, and Bengio}{2011}]{relu}
Glorot, X.; Bordes, A.; and Bengio, Y.
\newblock 2011.
\newblock Deep sparse rectifier neural networks.
\newblock In Gordon, G.; Dunson, D.; and Dudík, M., eds., {\em Proceedings of
  the Fourteenth International Conference on Artificial Intelligence and
  Statistics}, volume~15 of {\em Proceedings of Machine Learning Research},
  315--323.
\newblock Fort Lauderdale, FL, USA: PMLR.

\bibitem[\protect\citeauthoryear{Gneiting and
  Raftery}{2007}]{doi:10.1198/016214506000001437}
Gneiting, T., and Raftery, A.~E.
\newblock 2007.
\newblock Strictly proper scoring rules, prediction, and estimation.
\newblock {\em Journal of the American Statistical Association}
  102(477):359--378.

\bibitem[\protect\citeauthoryear{Haarnoja \bgroup et al\mbox.\egroup
  }{2017}]{energybased}
Haarnoja, T.; Tang, H.; Abbeel, P.; and Levine, S.
\newblock 2017.
\newblock Reinforcement learning with deep energy-based policies.
\newblock In {\em Proceedings of the 34th International Conference on Machine
  Learning, {ICML} 2017, Sydney, NSW, Australia, 6-11 August 2017},
  1352--1361.

\bibitem[\protect\citeauthoryear{Haarnoja \bgroup et al\mbox.\egroup
  }{2018a}]{latent}
Haarnoja, T.; Hartikainen, K.; Abbeel, P.; and Levine, S.
\newblock 2018a.
\newblock Latent space policies for hierarchical reinforcement learning.
\newblock In {\em Proceedings of the 35th International Conference on Machine
  Learning, {ICML} 2018, Stockholmsm{\"{a}}ssan, Stockholm, Sweden, July 10-15,
  2018},  1846--1855.

\bibitem[\protect\citeauthoryear{Haarnoja \bgroup et al\mbox.\egroup
  }{2018b}]{soft1}
Haarnoja, T.; Zhou, A.; Abbeel, P.; and Levine, S.
\newblock 2018b.
\newblock Soft actor-critic: Off-policy maximum entropy deep reinforcement
  learning with a stochastic actor.
\newblock In {\em Proceedings of the 35th International Conference on Machine
  Learning, {ICML} 2018, Stockholmsm{\"{a}}ssan, Stockholm, Sweden, July 10-15,
  2018},  1856--1865.

\bibitem[\protect\citeauthoryear{Haarnoja \bgroup et al\mbox.\egroup
  }{2018c}]{soft2}
Haarnoja, T.; Zhou, A.; Hartikainen, K.; Tucker, G.; Ha, S.; Tan, J.; Kumar,
  V.; Zhu, H.; Gupta, A.; Abbeel, P.; and Levine, S.
\newblock 2018c.
\newblock Soft actor-critic algorithms and applications.
\newblock {\em CoRR} abs/1812.05905.

\bibitem[\protect\citeauthoryear{Hallin \bgroup et al\mbox.\egroup
  }{2010}]{multivariate_qr2}
Hallin, M.; Paindaveine, D.; {\v{S}}iman, M.; Wei, Y.; Serfling, R.; Zuo, Y.;
  Kong, L.; and Mizera, I.
\newblock 2010.
\newblock Multivariate quantiles and multiple-output regression quantiles: From
  l 1 optimization to halfspace depth [with discussion and rejoinder].
\newblock {\em The Annals of Statistics}  635--703.

\bibitem[\protect\citeauthoryear{Henderson \bgroup et al\mbox.\egroup
  }{2018}]{drl_that_matters}
Henderson, P.; Islam, R.; Bachman, P.; Pineau, J.; Precup, D.; and Meger, D.
\newblock 2018.
\newblock Deep reinforcement learning that matters.
\newblock In {\em Proceedings of the Thirty-Second {AAAI} Conference on
  Artificial Intelligence, (AAAI-18), the 30th innovative Applications of
  Artificial Intelligence (IAAI-18), and the 8th {AAAI} Symposium on
  Educational Advances in Artificial Intelligence (EAAI-18), New Orleans,
  Louisiana, USA, February 2-7, 2018},  3207--3214.

\bibitem[\protect\citeauthoryear{Islam \bgroup et al\mbox.\egroup
  }{2017}]{reproducible}
Islam, R.; Henderson, P.; Gomrokchi, M.; and Precup, D.
\newblock 2017.
\newblock Reproducibility of benchmarked deep reinforcement learning tasks for
  continuous control.
\newblock {\em CoRR} abs/1708.04133.

\bibitem[\protect\citeauthoryear{Jones}{1992}]{Jones1992}
Jones, M.~C.
\newblock 1992.
\newblock Estimating densities, quantiles, quantile densities and density
  quantiles.
\newblock {\em Annals of the Institute of Statistical Mathematics}
  44(4):721--727.

\bibitem[\protect\citeauthoryear{Kingma and Welling}{2013}]{VAE}
Kingma, D.~P., and Welling, M.
\newblock 2013.
\newblock Auto-encoding variational bayes.
\newblock {\em arXiv preprint arXiv:1312.6114}.

\bibitem[\protect\citeauthoryear{Koenker and
  Hallock}{2001}]{quantile_regression_loss}
Koenker, R., and Hallock, K.
\newblock 2001.
\newblock Quantile regression: An introduction.
\newblock {\em Journal of Economic Perspectives} 15(4):43--56.

\bibitem[\protect\citeauthoryear{Koenker}{2005}]{quantile_regression}
Koenker, R.
\newblock 2005.
\newblock {\em Quantile Regression}.
\newblock Econometric Society Monographs. Cambridge University Press.

\bibitem[\protect\citeauthoryear{Lang}{2005}]{tanh_mono_net}
Lang, B.
\newblock 2005.
\newblock Monotonic multi-layer perceptron networks as universal approximators.
\newblock In Duch, W.; Kacprzyk, J.; Oja, E.; and Zadro{\.{z}}ny, S., eds.,
  {\em Artificial Neural Networks: Formal Models and Their Applications --
  ICANN 2005},  31--37.
\newblock Berlin, Heidelberg: Springer Berlin Heidelberg.

\bibitem[\protect\citeauthoryear{Lillicrap \bgroup et al\mbox.\egroup
  }{2015}]{DDPG}
Lillicrap, T.~P.; Hunt, J.~J.; Pritzel, A.; Heess, N.; Erez, T.; Tassa, Y.;
  Silver, D.; and Wierstra, D.
\newblock 2015.
\newblock Continuous control with deep reinforcement learning.
\newblock {\em arXiv preprint arXiv:1509.02971}.

\bibitem[\protect\citeauthoryear{Liu and Wang}{2016}]{SVGD}
Liu, Q., and Wang, D.
\newblock 2016.
\newblock Stein variational gradient descent: {A} general purpose bayesian
  inference algorithm.
\newblock In {\em Advances in Neural Information Processing Systems 29: Annual
  Conference on Neural Information Processing Systems 2016, December 5-10,
  2016, Barcelona, Spain},  2370--2378.

\bibitem[\protect\citeauthoryear{Matheson and Winkler}{1976}]{10.2307/2629907}
Matheson, J.~E., and Winkler, R.~L.
\newblock 1976.
\newblock Scoring rules for continuous probability distributions.
\newblock {\em Management Science} 22(10):1087--1096.

\bibitem[\protect\citeauthoryear{Minin and Lang}{2008}]{overview_mono_net_1}
Minin, A., and Lang, B.
\newblock 2008.
\newblock Comparison of neural networks incorporating partial monotonicity by
  structure.
\newblock In K{\r{u}}rkov{\'a}, V.; Neruda, R.; and Koutn{\'i}k, J., eds., {\em
  Artificial Neural Networks - ICANN 2008},  597--606.
\newblock Berlin, Heidelberg: Springer Berlin Heidelberg.

\bibitem[\protect\citeauthoryear{Mnih \bgroup et al\mbox.\egroup }{2015}]{DQN}
Mnih, V.; Kavukcuoglu, K.; Silver, D.; Rusu, A.~A.; Veness, J.; Bellemare,
  M.~G.; Graves, A.; Riedmiller, M.~A.; Fidjeland, A.; Ostrovski, G.; Petersen,
  S.; Beattie, C.; Sadik, A.; Antonoglou, I.; King, H.; Kumaran, D.; Wierstra,
  D.; Legg, S.; and Hassabis, D.
\newblock 2015.
\newblock Human-level control through deep reinforcement learning.
\newblock {\em Nature} 518(7540):529--533.

\bibitem[\protect\citeauthoryear{Mnih \bgroup et al\mbox.\egroup }{2016}]{A3C}
Mnih, V.; Badia, A.~P.; Mirza, M.; Graves, A.; Lillicrap, T.~P.; Harley, T.;
  Silver, D.; and Kavukcuoglu, K.
\newblock 2016.
\newblock Asynchronous methods for deep reinforcement learning.
\newblock In {\em Proceedings of the 33nd International Conference on Machine
  Learning, {ICML} 2016, New York City, NY, USA, June 19-24, 2016},
  1928--1937.

\bibitem[\protect\citeauthoryear{M{\"u}ller}{1997}]{muller1997integral}
M{\"u}ller, A.
\newblock 1997.
\newblock Integral probability metrics and their generating classes of
  functions.
\newblock {\em Advances in Applied Probability} 29(2):429--443.

\bibitem[\protect\citeauthoryear{Munos \bgroup et al\mbox.\egroup
  }{2016}]{Retrace}
Munos, R.; Stepleton, T.; Harutyunyan, A.; and Bellemare, M.~G.
\newblock 2016.
\newblock Safe and efficient off-policy reinforcement learning.
\newblock In {\em Advances in Neural Information Processing Systems 29: Annual
  Conference on Neural Information Processing Systems 2016, December 5-10,
  2016, Barcelona, Spain},  1046--1054.

\bibitem[\protect\citeauthoryear{Ostrovski, Dabney, and
  Munos}{2018}]{QN_generative_modeling}
Ostrovski, G.; Dabney, W.; and Munos, R.
\newblock 2018.
\newblock Autoregressive quantile networks for generative modeling.
\newblock In {\em Proceedings of the 35th International Conference on Machine
  Learning, {ICML} 2018, Stockholmsm{\"{a}}ssan, Stockholm, Sweden, July 10-15,
  2018},  3933--3942.

\bibitem[\protect\citeauthoryear{Parzen}{1979}]{parzen1979nonparametric}
Parzen, E.
\newblock 1979.
\newblock Nonparametric statistical data modeling.
\newblock {\em Journal of the American statistical association}
  74(365):105--121.

\bibitem[\protect\citeauthoryear{Richemond and Maginnis}{2017}]{WGF_}
Richemond, P.~H., and Maginnis, B.
\newblock 2017.
\newblock On wasserstein reinforcement learning and the fokker-planck equation.
\newblock {\em CoRR} abs/1712.07185.

\bibitem[\protect\citeauthoryear{Schulman, Abbeel, and Chen}{2017}]{q=ac}
Schulman, J.; Abbeel, P.; and Chen, X.
\newblock 2017.
\newblock Equivalence between policy gradients and soft q-learning.
\newblock {\em CoRR} abs/1704.06440.

\bibitem[\protect\citeauthoryear{Schulman \bgroup et al\mbox.\egroup
  }{2015}]{TRPO}
Schulman, J.; Levine, S.; Moritz, P.; Jordan, M.~I.; and Abbeel, P.
\newblock 2015.
\newblock Trust region policy optimization.
\newblock {\em CoRR} abs/1502.05477.

\bibitem[\protect\citeauthoryear{Schulman \bgroup et al\mbox.\egroup
  }{2016}]{GAE}
Schulman, J.; Moritz, P.; Levine, S.; Jordan, M.~I.; and Abbeel, P.
\newblock 2016.
\newblock High-dimensional continuous control using generalized advantage
  estimation.
\newblock In {\em 4th International Conference on Learning Representations,
  {ICLR} 2016, San Juan, Puerto Rico, May 2-4, 2016, Conference Track
  Proceedings}.

\bibitem[\protect\citeauthoryear{Schulman \bgroup et al\mbox.\egroup
  }{2017}]{PPO}
Schulman, J.; Wolski, F.; Dhariwal, P.; Radford, A.; and Klimov, O.
\newblock 2017.
\newblock Proximal policy optimization algorithms.
\newblock {\em CoRR} abs/1707.06347.

\bibitem[\protect\citeauthoryear{Sill}{1998}]{monotonic_net}
Sill, J.
\newblock 1998.
\newblock Monotonic networks.
\newblock In {\em Advances in neural information processing systems},
  661--667.

\bibitem[\protect\citeauthoryear{Sutton and Barto}{2018}]{sutton}
Sutton, R.~S., and Barto, A.~G.
\newblock 2018.
\newblock {\em Reinforcement Learning: An Introduction}.
\newblock The MIT Press, second edition.

\bibitem[\protect\citeauthoryear{Sutton \bgroup et al\mbox.\egroup
  }{2000}]{sutton2000policy}
Sutton, R.~S.; McAllester, D.~A.; Singh, S.~P.; and Mansour, Y.
\newblock 2000.
\newblock Policy gradient methods for reinforcement learning with function
  approximation.
\newblock In {\em Advances in neural information processing systems},
  1057--1063.

\bibitem[\protect\citeauthoryear{Tang and Agrawal}{2018}]{boostnormflow}
Tang, Y., and Agrawal, S.
\newblock 2018.
\newblock Boosting trust region policy optimization by normalizing flows
  policy.
\newblock {\em CoRR} abs/1809.10326.

\bibitem[\protect\citeauthoryear{Tarca, Grandjean, and
  Larachi}{2004}]{genetic_mono_net}
Tarca, L.~A.; Grandjean, B.~P.; and Larachi, F.
\newblock 2004.
\newblock Embedding monotonicity and concavity in the training of neural
  networks by means of genetic algorithms: Application to multiphase flow.
\newblock {\em Computers and Chemical Engineering} 28(9):1701 -- 1713.

\bibitem[\protect\citeauthoryear{{Todorov}, {Erez}, and {Tassa}}{2012}]{mujoco}
{Todorov}, E.; {Erez}, T.; and {Tassa}, Y.
\newblock 2012.
\newblock Mujoco: A physics engine for model-based control.
\newblock In {\em 2012 IEEE/RSJ International Conference on Intelligent Robots
  and Systems},  5026--5033.

\bibitem[\protect\citeauthoryear{Tukey}{1965}]{tukey1965part}
Tukey, J.~W.
\newblock 1965.
\newblock Which part of the sample contains the information?
\newblock {\em Proceedings of the National Academy of Sciences} 53(1):127--134.

\bibitem[\protect\citeauthoryear{Velikova, Daniels, and
  Feelders}{2006}]{mixture_mono_net}
Velikova, M.; Daniels, H.; and Feelders, A.~J.
\newblock 2006.
\newblock Mixtures of monotone networks for prediction.

\bibitem[\protect\citeauthoryear{Watkins}{1989}]{watkins1989learning}
Watkins, C. J. C.~H.
\newblock 1989.
\newblock {\em Learning from delayed rewards}.
\newblock Ph.D. Dissertation, King's College, Cambridge.

\bibitem[\protect\citeauthoryear{Wu \bgroup et al\mbox.\egroup }{2017}]{ACKTR}
Wu, Y.; Mansimov, E.; Liao, S.; Grosse, R.~B.; and Ba, J.
\newblock 2017.
\newblock Scalable trust-region method for deep reinforcement learning using
  kronecker-factored approximation.
\newblock {\em CoRR} abs/1708.05144.

\bibitem[\protect\citeauthoryear{You \bgroup et al\mbox.\egroup
  }{2017}]{lattice_mono_net}
You, S.; Ding, D.; Canini, K.; Pfeifer, J.; and Gupta, M.
\newblock 2017.
\newblock Deep lattice networks and partial monotonic functions.
\newblock In Guyon, I.; Luxburg, U.~V.; Bengio, S.; Wallach, H.; Fergus, R.;
  Vishwanathan, S.; and Garnett, R., eds., {\em Advances in Neural Information
  Processing Systems 30}. Curran Associates, Inc.
\newblock  2981--2989.

\bibitem[\protect\citeauthoryear{Zhang \bgroup et al\mbox.\egroup }{2018}]{WGF}
Zhang, R.; Chen, C.; Li, C.; and Carin, L.
\newblock 2018.
\newblock Policy optimization as wasserstein gradient flows.
\newblock In {\em Proceedings of the 35th International Conference on Machine
  Learning, {ICML} 2018, Stockholmsm{\"{a}}ssan, Stockholm, Sweden, July 10-15,
  2018},  5741--5750.

\end{thebibliography}

\clearpage
\appendix
\setcounter{secnumdepth}{1} 
\section{Likelihood in Quantile Functions}
\label{app:likelihood}
While there are different ways to approximate an arbitrary probability distribution $Y$, we want to highlight a useful trait of approximating the quantile function.
Specifically, we note that the partial derivative of the quantile function $F^{-1}_Y(\tau)$ with respect to the quantile $\tau$ - also known as the \emph{sparsity function} \cite{tukey1965part} or \emph{quantile-density function} \cite{parzen1979nonparametric} - has the interesting property~\cite{Jones1992}:
\[\frac{\delta}{\delta\tau}F^-1_Y(\tau) = \frac{1}{p_Y(F^{-1}_Y(\tau))}\]
where $p_Y(\cdot)$ is the probability density function of the distribution $Y$. In our setup, this allows us to retrieve the likelihood of an action $a_\tau = \hat{G}_\theta(\tau)$ as
\begin{equation*}
    \pi_\theta(a_\tau) = \frac{1}{\frac{\delta}{\delta\tau}\hat{G}_\theta(\tau)}
\end{equation*}
where $\pi_\theta$ is the probability density function of the policy distribution implicitly defined in our quantile function approximation $\hat{G}_\theta$.

\section{Monotonic Neural Networks}
\label{app:monotonic_net}
While there are several methods for monotonic function approximation in the literature \cite{monotonic_overview}, (partially) monotonic neural networks \cite{overview_mono_net_1,overview_mono_net_2} are mainly constructed using either the max-pool, min-pool architecture introduced by \citeauthor{monotonic_net}~\shortcite{monotonic_net} or a positive weight network with \emph{tanh}-activations as introduced by \citeauthor{tanh_mono_net}~\shortcite{tanh_mono_net}. Other approaches include training a monotonic network through genetic algorithms \cite{genetic_mono_net} or using a mixture of monotonic networks for prediction \cite{mixture_mono_net}. More recently, \citeauthor{lattice_mono_net}~\shortcite{lattice_mono_net} introduced a monotonic neural network with multidimensional lattices as non-linearities. We believe that this approach over-complicates the problem of monotonic function approximation with a neural network and restrain to a novel, simple yet effective way.

First of all, we note that a neural network is a combination of functions, most commonly a combination of linear embeddings (i.e., matrix multiplication) and element-wise non-linearities (e.g., tanh or ReLU). The easiest way to restrict a neural network to approximate only monotonic functions is therefore to restrict all constituent functions to be monotonic with respect to their inputs. This can be done by restricting the matrix multiplications to have only positive weights and use monotonic element-wise non-linearities, as was done with \emph{tanh}-activation by \citeauthor{tanh_mono_net}~\shortcite{tanh_mono_net}. 
Weight positivity is easily achieved by setting all parameters of the network weight matrices $\theta^W_{net}$ to the exponent of unconstrained weights $\theta^W$, i.e., $\theta^W_{net} = \exp(\theta^W)$ with $\exp(\cdot)$ applied element-wise, as was done by~\citeauthor{monotonic_net}~\shortcite{monotonic_net}. We can then optimize $\theta^W$ using standard stochastic gradient decent methods. In our experiments we initialize $\theta^W$ to
$\log\left(\mathcal{U}\left(\left[0,\sqrt{\frac{\sigma}{F_{in}}}\right]\right)\right)$
where $F_{in}$ is the number of neurons in the previous layer and $\sigma$ is a hyperparameter we set to 3. Note that this positivity constraint through exponentiation is only done for the weights and not for the biases, since the biases do not influence the monotinicity.

The more interesting architecture choice is the choice of non-linearity. In our work, we take advantage of the good performance reported for ReLU non-linearities \cite{relu}. Note that ReLU activation, $y = \max(0, x)$, is monotone with respect to its input. More so, it is convex, which would lead to a network only capable of approximating convex functions, if we restrict the network weights to be positive. To overcome this limitation, we use an adjusted ReLU activation on the hidden layers of our monotonic neural network. That is, we take half of the embedding dimensions and apply a ReLU activation, while we apply an inverse ReLU activation, $y = \min(0, x)$, to the remaining embedding dimensions. Note that the inverse ReLU activation is concave and a following linear layer with positive weight matrix can therefore make any combination of concave and convex functions, giving it the capability to approximate any monotonic function for a large enough hidden embedding.

\begin{table*}
    \centering
    \begin{tabular}{c|c|c|c}
         & Gaussian & Bi-Modal-Gaussian & Discontinuous Uniform \\ \hline
         Max-Min & \begin{tabular}{@{}c@{}}$0.048\pm 0.016$ \\ \small{lr = 0.01}\end{tabular} & \begin{tabular}{@{}c@{}}$0.031\pm 0.013$ \\ \small{lr = 0.1}\end{tabular} & \begin{tabular}{@{}c@{}}$\textbf{0.006}\pm \textbf{0.003}$ \\ \small{lr = 0.1}\end{tabular} \\ \hline
         tanh & \begin{tabular}{@{}c@{}}$0.055\pm 0.015$ \\ \small{lr = 0.01}\end{tabular} & \begin{tabular}{@{}c@{}}$0.028\pm 0.005$ \\ \small{lr = 0.0001}\end{tabular} & \begin{tabular}{@{}c@{}}$0.023\pm 0.018$ \\ \small{lr = 0.01}\end{tabular} \\ \hline
         ReLU & \begin{tabular}{@{}c@{}}$\textbf{0.028}\pm \textbf{0.009}$ \\ \small{lr = 0.01}\end{tabular} & \begin{tabular}{@{}c@{}}$\textbf{0.019}\pm \textbf{0.011}$ \\ \small{lr = 0.001}\end{tabular} & \begin{tabular}{@{}c@{}}$0.050\pm 0.009$ \\ \small{lr = 0.001}\end{tabular} \\
    \end{tabular}
    \caption{Comparison of different monotonic network architectures mean squared error after fitting different distributions with quantile regression. Shown is the mean and standard deviation of 5 random seeds, as well as the learning rate with which the result was achieved.}
    \label{tab:mono_nets}
\end{table*}

To compare our architecture against the max-pool/min-pool architecture of~\citeauthor{monotonic_net}~\shortcite{monotonic_net} and an architecture with \emph{tanh}-non-linearity as proposed by~\citeauthor{tanh_mono_net}~\shortcite{tanh_mono_net}, we train simple one hidden layer networks of each architecture with matched number of parameters (64 hidden units for the \emph{ReLU} and \emph{tanh} architecture, 96 hidden units for the \emph{Max-Min} architecture) to approximate the quantile function of some toy distributions based on a scalar quantile input $\tau$. $\tau$ is fed to the network as one dimensional input, sampled from a (scaled and shifted) uniform distribution over [-1, 1]. As 1-dimensional target distributions we choose a normal Gaussian (mean 0 and standard deviation 1), a bi-modal gaussian distribution (two Gaussians with mean -1 and 1 and both with standard deviation 0.5; a sample is drawn with equal probability from either of the two) as well as a discontinuous uniform distribution (samples are drawn with equal probability either uniformly from [-1,-0.5] or uniformly from [0.5, 1]). We train each network on 10,000 mini-batches of 128 samples each on the quantile regression loss~(\ref{expected_quantile_loss}). The networks are trained on the learning rates \{0.00001, 0.0001, 0.001, 0.01, 0.1\} and the best results achieved (averaged over 5 random seeds) are compared in Table~\ref{tab:mono_nets}.\footnote{Code to re-run this experiment is available in the supplementary material.}

\begin{figure}
    \centering
    \input{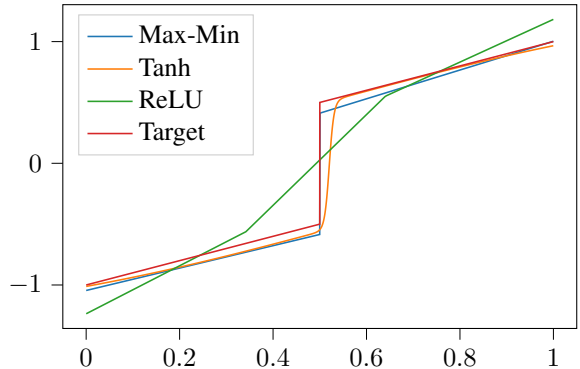}
    \caption{Learned quantile function approximations $\hat{G}_\theta(\tau)$ of the different monotonic network architectures to the quantile function of the discontinuous uniform distribution.}
    \label{fig:uniform_discontinuous}
\end{figure}

We note that while no architecture is significantly better than the others, we see a tendency of our \emph{ReLU} architecture to improve in normal distribution approximation. 
Further, note that the \emph{Max-Min} architecture requires comparatively large learning rates. We believe this is due to the fact that only a few parameters are updated with every sample (due to the pooling), therefore requiring a larger learning rate to be competitive with the other architectures. Lastly, we further investigate the bad performance of our ReLU architecture on the discontinuous uniform distribution task. In Figure~\ref{fig:uniform_discontinuous} we see that the ReLU network has difficulties to approximate the discontinuity. We believe that the stochastic nature of the task prohibits any weight in the network to grow large enough to approximate the discontinuity with a steep slope within the limited training time. Note that the other architectures provide inductive biases that help to deal with the discontinuity.

Nevertheless, we chose our ReLU architecture for our reinforcement learning agent, since we believe that the ability to approximate (multi-modal) normal distributions well out-weights the ability to approximate discontinuities well. Note that we might get even better results in our toy experiments if we had chosen otherwise.


\section{Constant Action Likelihood $\mu(a)$}
\label{app:constant_mu}
In our derivation in Section~\ref{qrrl} we initially assume the behaviour policy $\mu$ to be uniform, such that the action likelihood $\mu(a)$ gets absorbed as constant in the optimization. Later we then set $\mu\equiv\pi_{\theta_k}$, which is clearly a stretch on the assumption of uniformity, especially if we try to approximate multi-modal distributions with $\pi_{\theta_k}$. However, in early experiments we found that $\pi_{\theta_k}$ stays often close to a uniform distribution and mainly adjusts the support. Assuming the distribution goes from a brought, stochastic to a less stochastic distribution with smaller support, then the corresponding (constant) likelihood increases accordingly. This translates in a decreased learning rate over time (as the objective is multiplied by an ever smaller constant), which might also help the algorithm converge. We therefore retained the constant action likelihood approximation $\mu(a)\approx\text{const.}$ . Note however, that theoretically one can calculate the action likelihood based on the gradient of the action with respect to the quantile (see Appendix~\ref{app:likelihood}), store it alongside the action in the rollout data and correct the optimization objective accordingly. We leave the investigation of such a correction to future work, as action likelihoods in continuous action spaces can be non-trivial to handle.

\section{Pseudo Code}
\label{app:pseudocode}
A pseudo code version of our proposed algorithm can be seen in Algorithm~\ref{alg:pseudocode}.

\begin{algorithm}
\caption{Learning Policies through Quantile Regression}
\label{alg:pseudocode}
\begin{algorithmic}
\Require{Batch size $B$, number of mini-epochs $E$, generalized advantage estimation $\lambda$, discount factor $\gamma$, inner sample number $K$, randomly initialized policy network $\bm{\hat{G}}_\theta$}\\

\For{$k = 0,1,2,\dots$}

Gather $B$ time steps of experience by taking actions\\
\qquad\quad$\va_t = \bm{\hat{G}}_\theta(\bm{\tau},\vs_t)$ with $\bm{\tau}\sim\mathcal{U}([0,1]^d)$\\
\qquad Calculate the generalized advantage estimates\\ \qquad\quad$A_{\theta_k}^{GAE(\gamma,\lambda)}$ and normalize them over the batch
\For{$E$ epochs}\\
\qquad\qquad Sample a mini-batch of ($\vs_i,\va_i,A_{\theta_k,i}$) tuples\\
\qquad\qquad Sample $K$ $\bm{\tau}\sim\mathcal{U}([0,1]^d)$ for every tuple\\
\qquad\qquad Calculate $\mathcal{L}_\theta^k$ as mean over the batch and $\bm{\tau}$s\\
\qquad\qquad\quad based on equation~(\ref{final_objective})\\
\qquad\qquad Update $\theta$ to minimize $\mathcal{L}_\theta^k$ using Adam
\EndFor
\EndFor
\end{algorithmic}
\end{algorithm}

\section{Rock-Paper-Scissors}
\label{app:rps}
In our Rock-Paper-Scissors experiment we compare a Gaussian policy network against a quantile policy network. Both policies are trained as described in the main text against a Gaussian countering policy with the same specifications as the other Gaussian policy network detailed hereafter. The policies opposing each other in a game choose simultaneously an action and the winner is determined as follows:
\begin{itemize}
    \item If both policies chose a valid action, i.e., an action within one of the three action ranges depicted in Figure~\ref{fig:rps_action_space}, the winner is determined based on Rock-Paper-Scissors rules. That is, Rock wins over Scissors, Paper over Rock and Scissors over paper. If both policies chose the same action, the game results in a draw ($r=0$)
    \item If one of the policies chose an invalid action, i.e., an action outside the three action ranges, the other policy wins with any valid action.
    \item If both policies chose an invalid action, the game results in a draw.
\end{itemize}

As Gaussian policy network, we implemented a simple 2 layer fully connected neural network, which takes the two actions played in the last game as input, projects them to a hidden layer with 64 neurons and ReLU activation, and then back to a 2-dimensional embedding, from which the first dimension is taken as mean and the second dimension as log-standard deviation of the Gaussian defining the action distribution. 

Our quantile policy network also consists of 2 fully connected layers, where the first takes as input the one dimensional (scaled and shifted) quantile, sampled from a uniform distribution over [-1, 1], and projects it through a positive weight matrix to the hidden representation with 64 neurons. Half of the neurons in the hidden representation have ReLU activation, while the other half goes through an inverse ReLU non-linearity, as described in Appendix~\ref{app:monotonic_net}. The hidden representation is then projected with a positive weight matrix to a single output dimension, directly representing the action.
Both, Gaussian and quantile network, have trainable, unconstrained bias terms added in each layer. Also, we did not use multiple samples for the quantile loss in this experiment, i.e., we set $K=1$.

Note that the Gaussian network has (i) more parameters than the quantile network and (ii) more information given to it through the input: The Gaussian network can base its action on the last action played, giving it the ability to exploit any deterministic policy. Nevertheless, the quantile network can better fit the action space, leading to better results (see main text).

\section{Reinforcement Learning Experiments}
\label{app:RL}

For our reinforcement learning experiments, the toy Choice game as well as the MuJoCo experiments, we adjusted the PPO implementation published in the OpenAI baselines github repository: https://github.com/openai/baselines

That is, our policy network consists of a state feature extraction part equivalent to the PPO network architecture before the final projection to the Gaussian parameters. We replace this final projection through a quantile net per action dimension similar to the one described in Appendix~\ref{app:rps} with the only difference that we add the extracted features to the hidden representations of the quantile networks. We summarize the hyperparameters used in Table~\ref{tab:hypers}. For our baselines, we used the implementations provided by the authors.

\begin{table}[t]
    \centering
    \begin{tabular}{r|l}
         \textbf{Hyperparameter} & \textbf{Value} \\
         \hline
         Initial learning rate & 3e-4\\
         Learning rate schedule & linear decay \\
         Optimizer & Adam\\
         Adam epsilon & 1e-5 \\
         Discount factor & 0.99 \\
         GAE-lambda & 0.95 \\
         Steps taken per update & 2048 \\
         Number of epochs in each update & 10 \\
         Batch size & 32 \\
         Quantile loss samples K & 128 \\
         Temperature $\beta$ & 2
    \end{tabular}
    \caption{Hyperparameters used in the reinforcement learning experiments.}
    \label{tab:hypers}
\end{table}


\end{document}